\documentclass[letterpaper, 10 pt, conference]{ieeeconf} 
\IEEEoverridecommandlockouts 
\makeatletter
\def\endthebibliography{%
	\def\@noitemerr{\@latex@warning{Empty `thebibliography' environment}}%
	\endlist
}
\makeatother
\usepackage{graphicx}
\usepackage{amssymb}
\usepackage{amsmath}
\usepackage{setspace}
\usepackage{ulem}

\usepackage[dvipsnames,table]{xcolor} 

\usepackage{algorithmic}
\usepackage{algorithm}

\graphicspath{{./figures/}}

\title{\LARGE \bf Simple flagellated soft robot for locomotion near air-fluid interface}

\author{Yayun Du$^{1}$,
Andrew Miller$^{1}$,
Mohammad Khalid Jawed$^{1}$ 
\thanks{$^{1}$Department of Mechanical \& Aerospace Engineering, University of California, Los Angeles, 420 Westwood Plaza, Los Angeles, CA 90095}%
}

\begin{document}

\maketitle

\begin{abstract}
A wide range of microorganisms, e.g. bacteria, propel themselves by rotation of soft helical tails, also known as flagella. Due to the small size of these organisms, viscous forces overwhelm inertial effects and the flow is at low Reynolds number. In this fluid-structure problem, a competition between elastic forces and hydrodynamic (viscous) forces leads to a net propulsive force forward. A thorough understanding of this highly coupled fluid-structure interaction problem can not only help us better understand biological propulsion but also help us design bio-inspired functional robots with applications in oil spill cleanup, water quality monitoring, and infrastructure inspection. Here, we introduce arguably the simplest soft robot with a single binary control signal, which is capable of moving along an arbitrary 2D trajectory near air-fluid interface and at the interface between two fluids. The robot exploits the variation in viscosity to move along the prescribed trajectory. Our analysis of this newly introduced soft robot consists of three main components. First, we fabricate this simple robot and use it as an experimental testbed. Second, a discrete differential geometry-based modeling framework is used for simulation of the robot. Upon validation of the simulation tool, the third part of this study employs the simulations to develop a control scheme with a single binary input to make the robot follow any prescribed path.
\end{abstract}

\section{Introduction}
Inspired by the inherent structural compliance of living creatures, soft swimming robots are designed to be lifelike and better emulate the movement of creatures in nature. Such soft robots often exploit structural deformation for functionality. Propulsion of bacteria by rotation of flexible tail-like \textit{flagella}~\cite{silverman1977bacterial} is a source of inspiration for soft robot design. Flagella-propelled bacteria have been cited to be the ``most efficient machines in the universe''~\cite{dembski2008design} as they can swim at speeds up to tens of body lengths per second. Interestingly, large deformation and buckling in flagella can be used to control the swimming direction of bacteria~\cite{son2013bacteria}. In this paper, we adopt this paradigm of using deformation for functionality in soft structures.

\begin{figure}[h!]
	\centering
	\includegraphics[width=1\columnwidth]{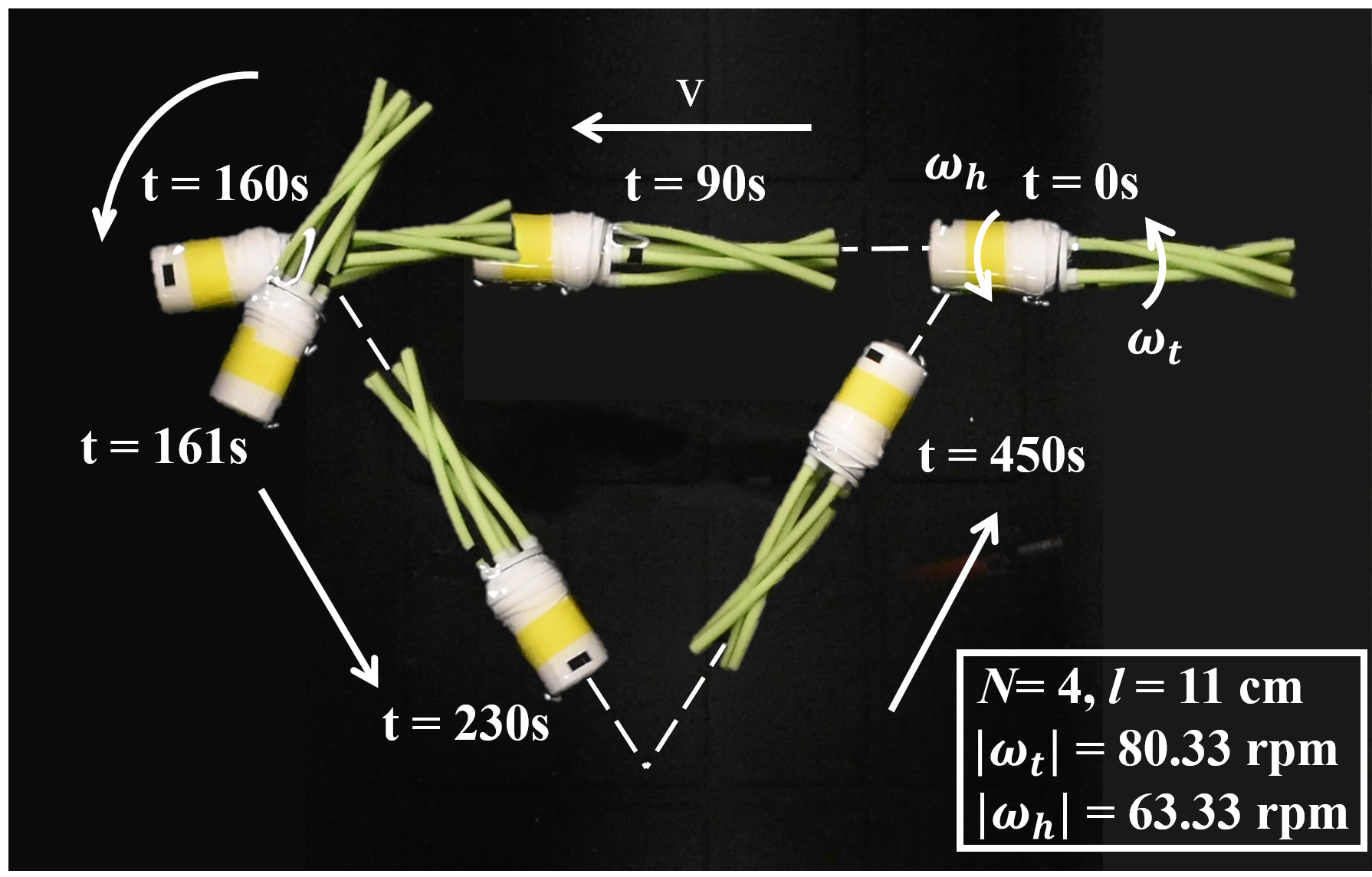}
	\caption{
	Snapshots of the robot (top view) moving along a triangular trajectory. Number of tails $N=4$; tail length $l=11$ cm; angular velocity of tail and head is $80.33$ and $63.33$rpm, respectively. The sign of the angular velocity is flipped at specific timepoints to achieve the triangular trajectory. Trajectory design is discussed in Section~\ref{sec:robotcontrol}.
	}
	\label{fig:overview}
\end{figure}

The typical fluid flow around a swimming bacterium is of low Reynolds number, around $10^{-4}$, where the viscous force dominates the inertial counterpart owing to the small size of bacterial cells. Scallop theorem~\cite{temel2013characterization} establishes that a motion invariant under time reversal cannot achieve net propulsion in this regime. Flagellar propulsion is a mechanism that overcomes this barrier. A flagellar bacterium consists of a cell body and one or more flagella; a rotary motor generates relative rotation between the cell body and the flagella. This rotation creates a net propulsive force forward. Typically, bacterial flagella are helical and their propulsion has been extensively studied~\cite{lauga2009hydrodynamics}. In these studies, the viscosity and density of the fluid medium are usually assumed to be constant. However, in practice, such ideal fluid medium is not feasible. Near the boundary between fluid and air (or at the interface between two immiscible fluids), viscosity and density both vary spatially. We will show that this variation in viscosity can be exploited to build a very simple robot (composed of naturally \texttt{straight} flagella) that is capable of following any prescribed trajectory near the boundary. 
Fig.~\ref{fig:overview} shows snapshots of the robot moving along a triangular trajectory. This simple low-cost robot with a single binary control input can have applications in ocean oil spill cleanup, water quality monitoring, and pipe inspection. Interestingly, it has been reported that the motion of flagellated bacteria near air-liquid interface is circular~\cite{lemelle2010counterclockwise}. If the angular velocity of the motor is constant in the robot introduced in this study, its trajectory is also circular.

Due to the simplicity of the robot, it is amenable to miniaturization. A variety of robots have been developed in microscale for propulsion in marine environments. Microscale mobile robot fabrication, such as artificial bacterial flagella~\cite{zhang2009artificial, kim2016fabrication}, is restricted by the key bottleneck: miniaturization of power source and onboard actuation. The corresponding control strategies are often dependent on external magnetic field. While our prototype robot is centimeter-sized, we use a viscous fluid medium (glycerin) to maintain low Reynolds number. The findings are mostly presented in non-dimensional form and do not depend on the size of the system (as long as the Reynolds number is low). In the future, the simplicity of the proposed robot design can be exploited to develop untethered autonomous micro-robots.



In this work, we develop an economical centimeter-scale, simple-to-assemble, and self-contained robot comprised of a cylindrical head and a rotating disk containing two or more soft polymeric tails, actuated by the motor within the head. The motor generates a relative rotation between the head and the tails; therefore, the robot head and tails rotate in opposite directions. The magnitude of the angular velocities are determined by the torque balance of the system. The rotation leads to hydrodynamic (viscous) forces on the soft tails leading to elastic deformation; this deformation generates a net propulsive force that is used by the robot to translate in fluid. If the robot is in an infinite fluid bath, the direction of motion is parallel to the axis of the cylindrical head. However, in practice, such fluid bath with uniform viscosity and density is not practical. We exploit this variation and the robot (under constant angular velocity) moves along a line that is slanted with the axis of the head. Depending on the sign of the angular velocity, the robot moves clockwise or anti-clockwise along a circle. By periodically switching the sign of the angular velocity, the robot achieves a net translation along a straight line. We show that the robot can move along a straight line simply by switching the angular velocity; a constant angular velocity lets the robot make a turn. A simple control law is designed where the robot approximates a prescribed trajectory by a piece-wise linear function. To understand the physical principles, a simulation tool is developed where the structure is modelled using the Discrete Elastic Rods (DER) algorithm~\cite{bergou2010discrete, jawed2018primer} and the fluid forces are implemented using Resistive Force Theory (RFT)~\cite{gray1955propulsion}. We show that elementary physics can be used to explain the propulsion mechanism of this robot.

Our contributions are as follows. We introduce a simple untethered soft robot that exploits variation in viscosity and elastic deformation in its tails to follow a pre-planned trajectory. A complete framework comprising of experiments, simulations, and controls is described to study the flagellated robot. The simulation tool is faster than real-time on a contemporary computer and can be used to generate data to formulate a control strategy. The physics behind the locomotion is elaborated. The simplicity of the robot and the small number of moving parts can eventually lead to miniaturization of this robot.

The remainder of the paper is organized as follows. We provide details on experiments and simulations in Section II. In Section III, we list the relevant physical parameters that affect the motion of the robot. Next, a simple control scheme that needs a single binary input for the robot to pursue the desired motion path is given in Section IV. Eventually, Section V concludes the paper.

\section{Methods}
\label{sec:method}

\begin{figure}[h!]
	\centering
	\includegraphics[width=1\columnwidth]{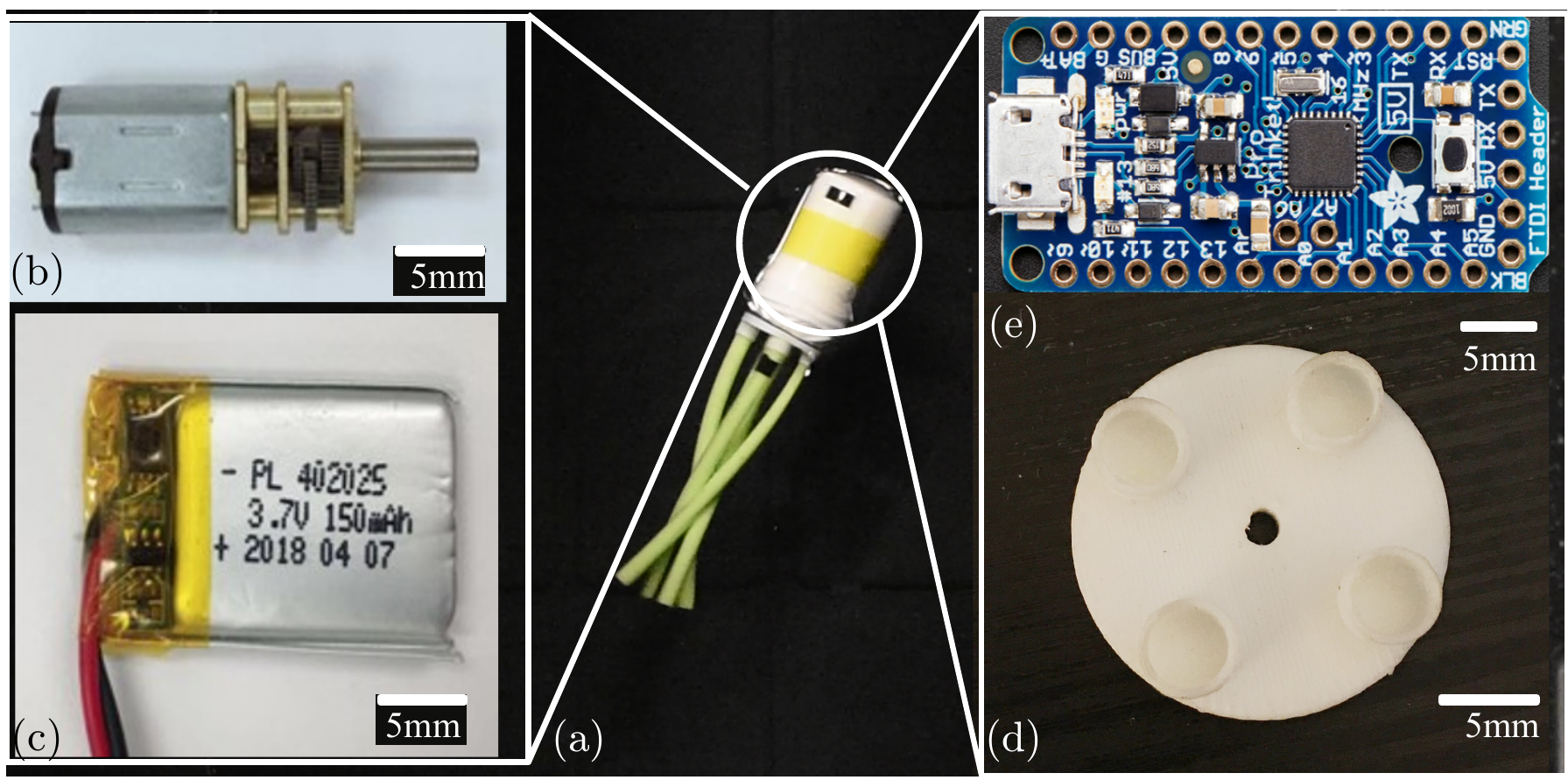}
	\caption{Compositive view of the experimental setup. (a) The robot with $n=4$ tails in glycerin (top view). The head is comprised of (b) a DC geared motor, (c) a battery, (d) a 3D printed circular disc connecting the tails to the rotating motor shaft, and (e) a microcontroller to control the rotational speed of the motor. 
	}
	\label{fig:experimentSetup}
\end{figure}

\begin{figure*}[t!]
	\centering
	\includegraphics[width=0.7\textwidth]{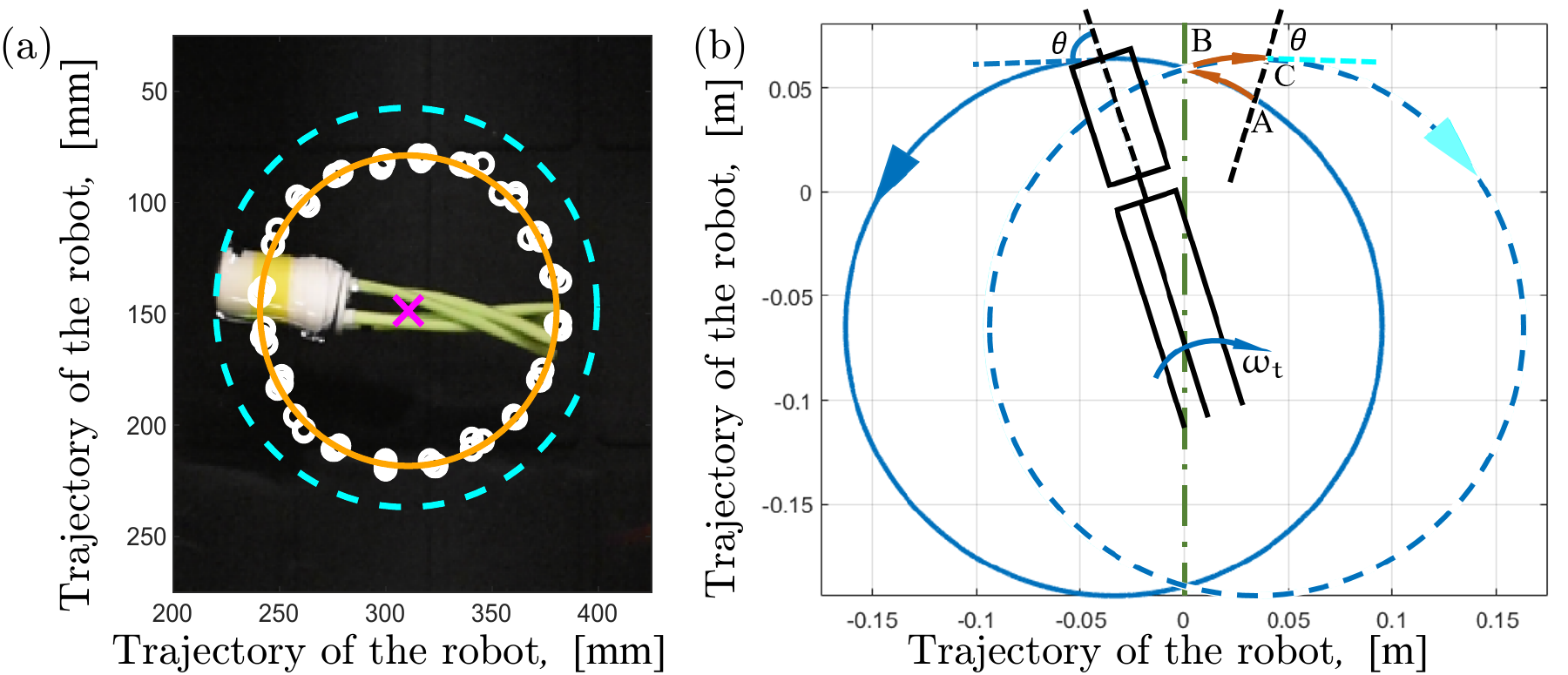}
	\caption{ (a) Circular movement of the robot in glycerin. 
	(b) Trajectories of the robot head (solid and dashed circles) when the robot starts from the same place but with different signs of the angular velocity of the motor.
	}
	\label{fig:robot}
\end{figure*} 


\subsection{Robot design and experimental setup}
 \label{subsec:experiment}
  

Glycerin with a density of $1.26$ g/mL and viscosity $\mu_0 = 1.49$ Pa-s at $25^\circ$C is selected as the fluid medium. The density of our lightweight and compact robot is slightly less than that of glycerin and it remains submerged near the air-fluid boundary. The robot in Fig.~\ref{fig:experimentSetup} is comprised of a head, multiple elastic tails, and a 3D-printed plate attached to the motor shaft to hold these tails. The robot head is a cylinder with a radius of $1.6$cm and height of $6$cm, which contains inside (b) one DC geared motor (uxcell) with $3$V nominal voltage, $0.35$W nominal power and $0.55$A stall current (c) one $3.7$V $200$mAh rechargeable 502025 LiPo batteries, and (e) a $5$V, $16$MHz adafruit pro trinket. The motor is embedded inside the head with its shaft protruding out, and its rotation direction and speed are controlled by changing the PWM value in the program running in the trinket. The radius of the cylindrical head is $R=1.6$cm. Some copper wires are attached to the outer surface of the robot head to make it balanced horizontally. During all experiments, the robot\rq{}s tails are fully submerged in glycerin while 30\% of the head is exposed to the air. In order to count the rotation speed of robot's head and tails clearly and conveniently, we stick a colored marker on one side of the robot's head and one of its tails. A digital camera (Nikon D3400) is used to record the robot\rq{}s movement from the bird\rq{}s eye view with its lens facing right down. The tails are made from Vinyl Polysiloxane using well established molding and casting techniques~\cite{jawed2015propulsion}. The Young\rq{}s modulus is $E=1.2$ MPa~\cite{jawed2015propulsion} and cross-sectional radius is $r_0=3.2$ mm. Since the material is near incompressible (Poisson ratio $\nu \approx 0.5$), the shear modulus is $G = E/3$. In order to generate enough of experimental data for parameter fitting in simulations, we vary the number of tails, $N = 2, 3, 4, 5$, and the length of tails, $l = 5, 7, 9, 11, 13, 15$ cm, with a DC geared motor mentioned above actuating the tails with a rated angular velocity of $150$ rpm. Note that the actual angular velocity of the motor varies depending on the number of tails and is not necessarily $150$ rpm, which ensures a Reynolds number $\textless 10^{-1}$.

\subsection{Experiment trials}
\label{subsec:experimentResult}

Images are extracted from the recorded experimental videos for data processing.  Fig.~\ref{fig:robot}(a) shows the trajectories of the tip of the robot head and the tip of a tail for a constant value of angular velocity of the motor ($\omega=143.66$ rpm). 
The rotation directions of the robot head and tails around the long axis (i.e. axis of the cylindrical head) are opposite, as the system is untethered and torque-balanced. 
If the magnitude of the angular velocities of the head and the tail are $\omega_h$ and $\omega_t$, respectively, and the angular velocity of the motor is $\omega$, then $| \omega_h | + | \omega_t | = | \omega |$.
The torque on the robot's head is balanced by the torque on the tails. 
As illustrated in Fig. \ref{fig:robot}(a), we also find that the whole robot circles around the vertical axis that is perpendicular to the air-fluid interface ($y$-axis in Fig. \ref{fig:dragOnHead}(a)) when its motor rotates unidirectionally, clockwise or counterclockwise. The open circle are the trajectory of the tip of robot tails; these points are fitted to the solid circle with the cross sign as the center. Similarly, the dashed circle is the circle fit to the trajectory of the tip of the robot head with the cross sign as the center.

Next, when we flip the sign of the angular velocity of the motor from the same initial orientation, the robot turns to circle around the vertical axis in the opposite direction. Specifically, as shown in Fig. \ref{fig:robot}(b), the solid and dashed circles are the trajectories of the robot head when the whole robot circles clockwise and counterclockwise (about the vertical $y$-axis), respectively. In both the cases, the initial orientation is along the dash-dot line. These two circles have the same radius but do not coincide with each other. To understand this, note that the angle, $\theta$, between the long axis of the robot (dashed line in Fig.~\ref{fig:robot}(b)) and the tangential direction of the circular trajectory is not $90^\circ$. As a result, if the tip of the robot's head starts to rotate from point $A$ and rotates counterclockwise along curve $AB$ first and then rotates along curve $BC$ after flipping the rotation direction of the motor, the robot will move forward and generate a translational movement. The net translation is the line segment $AC$. In summary, periodically switching the angular velocity $\omega$ of the robot between positive and negative values (keeping the same magnitude) results in a net straight-line trajectory. If the angular velocity of the robot about $y$-axis is $\omega_{yr}$, the robot will make a turn by an angle $\alpha$ if the motor\rq{}s angular velocity is maintained at $\omega$ for a period of $\alpha / \omega_{yr}$. Note that $\omega_{yr}$ is a function of various geometric, material, and fluid parameters (See Section~\ref{sec:parameterSweep}). This is where a comprehensive simulation tool and a physics-based understanding, to be discussed in the next section, can guide us to develop a control law.


\subsection{Numerical Simulation}
\label{subsec:simulation}

\begin{figure}[h!]
	\centering
	\includegraphics[width=1\columnwidth]{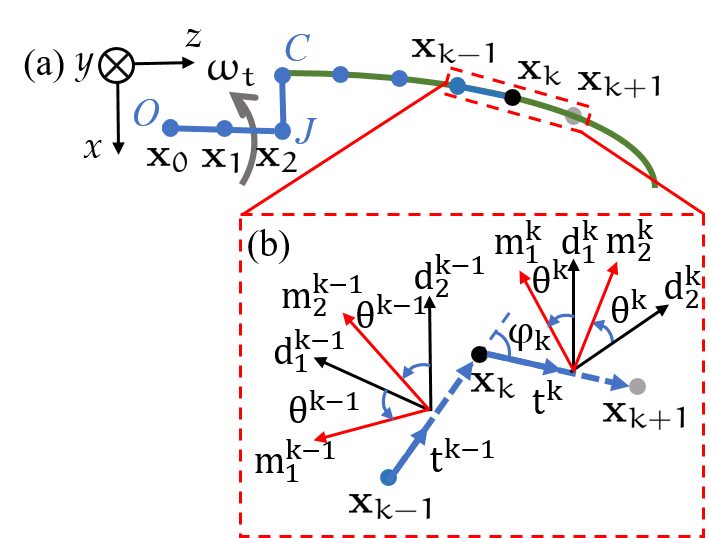}
	\caption{
	(a) Discrete representation of the soft robot.
	(b) Three nodes, two edges, and the associated reference and material frames.
	}
	\label{fig:DER_RFT}
\end{figure} 


\begin{figure*}[t!]
    \centering
    \includegraphics[width=0.6\textwidth]{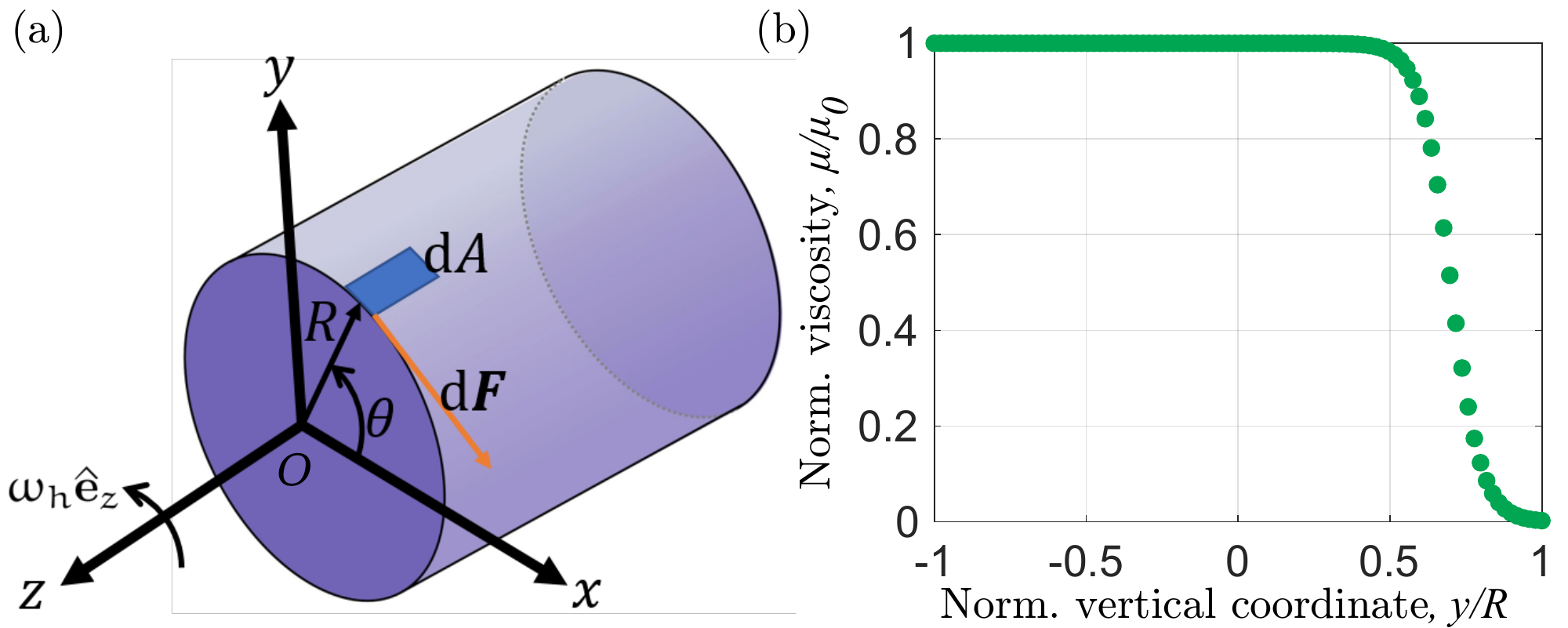}
    \caption{(a) Schematic showing drag $\mathrm d \mathbf{F}$ on the cylindrical head when the head is rotating along its long axis ($z$-axis). (b) Normalized viscosity as a function of normalized vertical coordinate $\frac{y}{R}$.}
    \label{fig:dragOnHead}
\end{figure*}

We develop a numerical simulation based on the Discrete Elastic Rods (DER) method; a tutorial exposition to DER can be found in Ref.~\cite{jawed2018primer}. In DER, the robot is discretized into $n$ nodes, as shown in Fig. \ref{fig:DER_RFT}(a). There are three nodes ($\mathbf x_0, \mathbf x_1,$ and $\mathbf x_2$) on the head and equal number of nodes on each tail (for illustration purposes, only one leg is shown in Fig.~\ref{fig:DER_RFT}(a)). It is necessary to have three nodes on the head to model actuation using a natural twist that varies with time (more on this later in this section). Two adjacent nodes, $\mathbf x_k$ and $\mathbf x_{k+1}$, are connected by an \textit{edge}, $\mathbf e^k = \mathbf x_{k+1} - \mathbf x_{k}$.
%
%
Each edge is associated with an orthonormal reference frame, $\{\mathbf t^k, \mathbf d_1^k, \mathbf d_2^k\}$, and an orthonormal material frame, $\{\mathbf t^k, \mathbf m_1^k, \mathbf m_2^k\}$. Both of these frames are \textit{adapted}, i.e. the first director $\mathbf t^k$ is the unit vector along the edge $\mathbf e^k$. The simulation moves forward with time taking small steps of $\Delta t$. During the simulation loop, the reference frame is updated through parallel transport in time. We omit the details of time parallel transport; Ref.~\cite{jawed2018primer} includes a pedagogical introduction to this method. Since the material frame shares a common director $\mathbf t^k$ with the reference frame, only a scalar angle $\theta^k$ (see Fig.\ref{fig:DER_RFT}(b)) is necessary to describe the material frame. The degrees of freedom (DOF) vector of the robot is then $\mathbf q = \left[ \mathbf x_0, \mathbf x_1, \ldots, \mathbf x_{n-1}, \theta^0, \theta^1, \ldots, \theta^{m-1} \right]$, where $n$ is the number of nodes and $m$ is the number of edges. The total number of DOF is $\texttt{ndof} = 3n + m$.

The core of the simulation is a solver (integrator) of following equations of motion.
\begin{equation}
m_i \ddot{q}_i = F^e_i + F^h_i,
\end{equation}
where $m_i$ is the lumped mass at the $i$-th DOF, $q_i$ is the $i$-th element of the DOF vector, $F^e_i$ is the $i$-th element of the $\texttt{ndof}$-sized elastic force vector $\mathbf F^e$, and $F^h_i$ is the $i$-th element of the $\texttt{ndof}$-sized external (hydrodynamic) force vector $\mathbf F^h$. Hereafter, dot $\dot{(\;)}$ represents derivative with respect to time.

First, we describe the elastic forces. The elastic energy is composed of three modes: stretching, bending, and twisting. Each component is given by 
\begin{equation}
    \begin{aligned}
        E_k^s &= \frac{1}{2} EA \left( \frac{\mathbf x_{k+1} - \mathbf x_k}{\bar{\mathbf e}^k} - 1 \right)^2 |\bar{\mathbf e}^k | \\
        E_k^b &= \frac{1}{2} EI (|\kappa_k - \kappa_k^0 |)^2 \frac{1}{l_k} \\
        E_k^t &= \frac{1}{2} GJ (|\tau_k - \tau_k^0 |)^2 \frac{1}{l_k}
    \end{aligned}
\end{equation}
where 
$EA = E \pi r_0^2$, $EI = \pi E r_0^4/4$, $GJ = \pi G r_0^4/2$, 
$| \bar{\mathbf e}_k |$ is the length of edge $\mathbf e_k$ in the undeformed state, $\kappa_k$ is the curvature vector at node $\mathbf x_k$ (related to the \textit{turning angle} $\phi_k$ in Fig.~\ref{fig:DER_RFT}(b)) while $\kappa_k^0$ is the undeformed curvature for the same node, $\tau_k$ is the integrated twist (related to $\theta^{k+1} - \theta^k$ in Fig.~\ref{fig:DER_RFT}(b)) while $\tau_k^0$ represents the natural twist  at node $\mathbf x_k$, and $l_k = (|\bar{\mathbf e}_{k-1}| + |\bar{\mathbf e}_k| )/ 2$ is the Voronoi length of the node in undeformed state. 
The total elastic energy is $E^e = \sum_k E_k^s + \sum_k E_k^b + \sum E_k^t$. The elastic force vector is simply $\mathbf F^e = - \frac{\partial}{\partial \mathbf q} E^e$.

It is important to note that the elastic stiffness parameters are not the same throughout the rod. These parameters for the soft tails are described in the previous section. However, as the head and disc are rigid ($OJC$ portion in Fig.~\ref{fig:DER_RFT}(a)), we set the values of $EA, EI, GJ$ on this segment to be very large so that no deformation takes place.

In order to mimic actuation by the motor rotating at an angular velocity $\omega(t)$, we set the natural twist of the second node ($\tau^0_1$) to be
\begin{equation}
    \tau_1^0(t) = \omega (t).
\end{equation}


Next, we describe the formulation of the hydrodynamic force (i.e. viscous drag) vector $\mathbf F^h$.

\textbf{Hydrodynamic force on robot head:}
The cylindrical head with radius $R$ is translating with a velocity $\dot{\mathbf x}_1$ and rotating about its axis with an angular velocity of $\omega_h \equiv \dot{\theta}^0$. The hydrodynamic drag on a cylinder (external force on $\mathbf x_1$ in DER) can be decomposed into two parts:
\begin{equation}
    \mathbf F = \mathbf F_v (\dot{\mathbf x}_1) + \mathbf F_{\omega} (\omega_h),
    \label{eq:headTranslationDrag}
\end{equation}
where $\mathbf F_v (\dot{\mathbf x}_1)$ and $\mathbf F_{\omega} (\omega_h)$ are the drag forces due to translation and rotation, respectively. The former quantity is a function of the translational velocity, $\dot{\mathbf x}_1$, of the head while the latter is a function of the angular velocity, $\omega_h$. 

Drag due to the translation on a sphere is given by Stokes' law as 
\begin{equation}
    \mathbf F_v = - 6\pi \mu_0 R \dot{\mathbf x}_1,
    \label{eq:stokeslaw}
\end{equation}
where $R$ is the radius of the spherical object and $\dot{\mathbf x}_1$ is the velocity of the object relative to the fluid. Since the robot head is cylindrical and there is no closed form expression for drag on a cylinder, we use a numerical coefficient $C_t$ (to be evaluated through data fitting) to express the drag as 
\begin{equation}
    \mathbf F_v = - C_t 6\pi \mu_0 R \, \dot{\mathbf x}_1.
    \label{eq:dragTranslation}
\end{equation}
For the robot studied in this paper, the viscosity varies along the vertical direction. Fig. \ref{fig:dragOnHead}(a) shows a schematic of the head and $x-y-z$ is the body fixed frame. The vertical direction $y$ is perpendicular to the air-fluid interface. This interface where the viscosity changes rapidly from $\mu_0$ (fluid) to $0$ (air) is at $y \sim R$. The fitting parameter $C_t$ in Eq.~\ref{eq:dragTranslation} also depends on the functional relationship between viscosity $\mu$ and vertical position $y$.

\begin{figure*}
	\centering
	\includegraphics[width=0.8\textwidth]{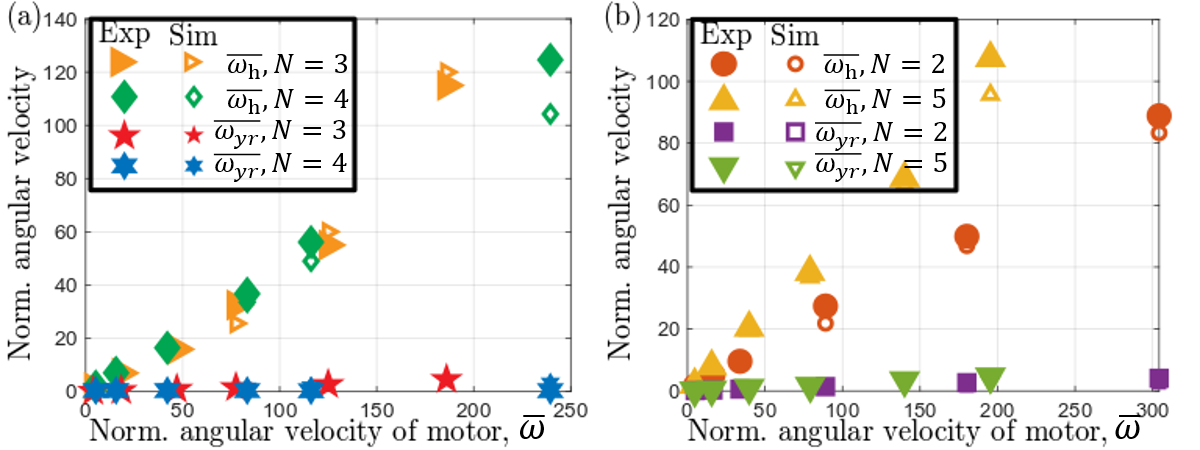}
	\caption{(a) Experimental and simulation data on $\bar \omega_h$ and $\bar \omega_{yr}$ as functions of the normalized angular velocity of the motor, $\bar \omega$, at two different values of the number of tails ($N = 3$ and $4$). This data are used to estimate $C_t, C_r, C_{yr}$.
	(b) Same data from experiments and simulations but with $N = 2$ and $5$. In simulations, the estimated values of $C_t, C_r, C_{yr}$ from (a) were used.
	}
	\label{fig:comparisonTailNum}
\end{figure*} 

Critical to the propulsion of this soft robot is the drag force $\mathbf F_{\omega}$ originating from this variation in viscosity.
%
The viscosity $\mu$ is a function of the y-coordinate, i.e. $\mu = \hat{\mu}(y)$. The specific functional form of $\mu$ does not matter as we will be using fitting parameters. We pick the following expression for viscosity,
\begin{equation}
    \mu = \mu_0 \frac{1}{1 + \exp \left( k\, \left( \frac{y - h}{R} \right) \right)},
\label{eq:viscosity}
\end{equation}
where $h$ is the location (close to the inter-medium boundary) where glycerin starts to mix with air and $k$ is the ``sharpness" of the transition from $\mu=\mu_0$ to $\mu=0$. In Fig.~\ref{fig:dragOnHead}, we used $h = 0.7 R$ and $k=20$. Note that Eq.~\ref{eq:viscosity} is an analytical approximation to the Heaviside function. 

Referring to Fig.~\ref{fig:dragOnHead}, a small area element $\mathrm{d}A = R \mathrm{d}\theta \mathrm{d}z$ on the surface of the cylinder rotating at an angular velocity of $\omega_h$ (along the $z$-axis) is picked. The magnitude of the force on this infinitesimal element is
\begin{equation}
    dF \sim \mu \omega_h \; R \mathrm{d}\theta \mathrm{d}z,
\end{equation}
with its direction along negative $\hat {\mathbf e_\theta}$, which is the unit vector along the tangential direction. The force along the $x$ axis is
\begin{equation}
    dF_x \sim dF \sin \theta = \mu \omega_h \sin \theta \; R \mathrm{d}\theta \mathrm{d}z,
    \label{eq:dFx}
\end{equation}
and the force along the $y$ axis is
\begin{equation}
    dF_y \sim - dF \cos \theta = - \mu \omega_h \cos \theta \; R \mathrm{d}\theta \mathrm{d}z.
    \label{eq:dFx}
\end{equation}

The horizontal component ($x$ axis) of the total force on the cylinder with length $L$ is obtained by integrating $dF_x$;
\begin{align}
    F_x  & \sim \int_{z=0}^{L} \int_{\theta=0}^{2\pi} \mu \omega_h \sin \theta \; R \mathrm{d}\theta \mathrm{d}z,\\
    \implies F_x = -1.403 \mu_0 \omega_h R L.
    \label{eq:numbericalIntegration}
\end{align}

Since we do not know the exact form of $\mu = \hat \mu (y)$,  a fitting parameter $C_{yr}$ is used and Eq. \ref{eq:numbericalIntegration} can be reformulated as 
\begin{equation}
    F_x = - C_{yr} \omega_h \mu_0 R L.
    \label{eq:F_x}
\end{equation}

The vertical component ($y$ axis) of the total force is
\begin{equation}
    F_y  \sim - \int_{z=0}^{L} \int_{\theta=0}^{2\pi} \mu \omega_h \cos \theta \; R \mathrm{d}\theta \mathrm{d}z = 0,
\end{equation}
i.e. there is no vertical hydrodynamic force.

In summary, the hydrodynamic drag on the head (applied on the center of mass of the head) due to rotation ($\omega_h$) is
\begin{equation}
\mathbf {F}_\omega (\omega_h) = - C_{yr} \omega_h \mu_0 R L \hat{\mathbf e_x}.
\end{equation}

The hydrodynamic moment on the head (applied on the first edge $\theta^0$ in DER) is 
\begin{equation}
    F_\omega = - C_{r} 8\pi \omega_h \mu_0 R^3,
    \label{eq:headRotationDrag}
\end{equation}
where $C_r$ is a numerical prefactor (fitting parameter in our study). Note that if the head was spherical, we would have $F_\omega = - 8\pi \omega_h \mu_0 R^3$.

\textbf{Hydrodynamic force on tails:}
The hydrodynamic force on the nodes belonging to the soft tails is formulated using RFT~\cite{gray1955propulsion, rodenborn2013propulsion}. The force on node $\mathbf x_k$ (moving with velocity $\dot{\mathbf x}_k$) is
%
\begin{equation}
    \mathbf F_\textrm{RFT} = 
    - \mu_{\parallel} \, (\mathbf{t} \cdot \dot{\mathbf x}_k )\mathbf{t} l_k - \mu_{\perp} [ \dot{\mathbf x}_k - (\mathbf{t}\cdot \dot{\mathbf x}_k )\mathbf t ] l_k , 
\label{eq:RFT}
\end{equation}
where $\mathbf t$ is the tangent vector on node $\mathbf x_k$, $l_k$ is the Voronoi length (described earlier), and $\mu_\parallel = 2\pi \mu_0/[\log(l/r_0)-\frac{1}{2}]$ and  $\mu_\perp = 4\pi \mu_0/[\log(l/r_0)+\frac{1}{2}]$ are the RFT drag coefficients along the tangential and perpendicular directions. 


The expressions of the forces in Eqs.~\ref{eq:headTranslationDrag}, ~\ref{eq:headRotationDrag}, and ~\ref{eq:RFT} are used to populate the external force vector $\mathbf F^h$ of size $\texttt{ndof}$.

\begin{figure*}[t!]
    \centering
    \includegraphics[width=0.75\textwidth]{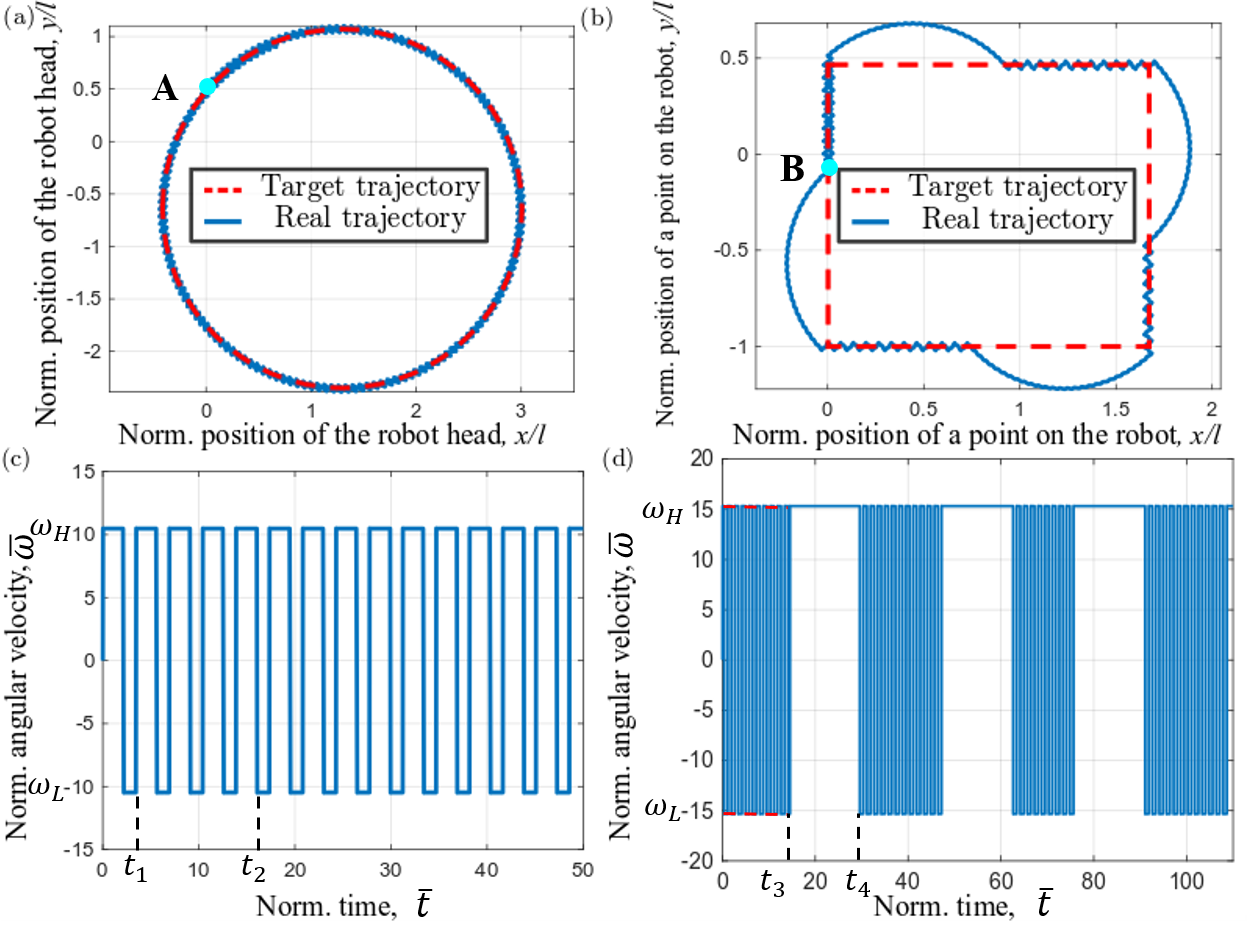}
    \caption{(a) Prescribed circular trajectory (dashed line) and real path (solid line) viewed from top. (b) Prescribed rectangular path (dashed line) and the real path of the robot (solid line). In (a) and (b), position has been normalized by tail length, $l$.
    Corresponding control signals (angular velocity) with time for (a) circular and (b) square trajectories.
    }
    %
    %
    %
    %
    %
    \label{fig:trajectory_design}
\end{figure*}

\textbf{Parameter fitting:}
As mentioned in Section \ref{subsec:experimentResult}, the tail length at each tail number varies from $5-15$ cm in experiments and we have 6 data-points for a specific tail number. Now that the hydrodynamic forces on the robot have been analyzed, there are three numerical prefactors ($C_t$, $C_r$, and $C_{yr}$) that need to be obtained from data fitting.
Our fitting strategy is to take the experimental data for $N=3$ and $N=4$ tails and find the set of parameters ($C_t$, $C_r$, and $C_{yr}$) that result in the best match between experiments and simulations. To evaluate the fitness of a given set of parameters, we use the following metrics: (i) angular velocity of the head, $\omega_h$ and (ii) angular velocity of the robot around the vertical axis, $\omega_{yr}$. In the experiments, we used the same motor with a full PWM value signal sent by the microcontroller. As we vary the length of the tails, the actual angular velocity of the motor, $\omega$, changes. Fig. \ref{fig:comparisonTailNum}(a) presents $\omega_h$ and $\omega_{yr}$ as functions of $\omega$. All other parameters (except $l$) are kept fixed.


The best fitting parameter set that realizes the smallest error, $14.8\%$, between experimental and simulation results in Fig. \ref{fig:comparisonTailNum}(a) is $C_t = 4.0 \pm 0.33, C_r = 2.06 \pm 0.156, C_{yr} = 6.0 \pm 0.5$. After the application of this fitting parameter set, the predicted simulation results for 2 and 5 tails turn out to match well with experiments with a 10\% error as shown in Fig. \ref{fig:comparisonTailNum}(b). This agreement indicates that the physics of this robot has been captured using the hydrodynamic model presented earlier in this section.


\section{Parameter space}
\label{sec:parameterSweep}

In this section, we list the relevant physical parameters that affect the motion of the robot. Note that there is an intrinsic time-scale~\cite{coq2008rotational} in this problem $\mu l^4/EI$. We use this time-scale to normalize various quantities (overbar represents normalization), e.g. $\bar \omega = \omega \mu l^4/EI$ is normalized angular velocity of the motor and $\bar t = t EI/[\mu l^4]$ is normalized time. The set of physical parameters that describe the system is $\{ C_t, C_r, C_{yr}, l/R, L/R, l/r_0, \bar{\omega}, N \}$; these are the inputs to our simulation tool. The angular velocity is a function of time. The simulation outputs the trajectory of the robot with time. In the next section, we will address the inverse problem where $\bar{\omega}$ has to be computed, given a prescribed trajectory.

The output of the simulation (i.e. trajectory of the robot) when $\bar{\omega}$ is constant with time can be encapsulated with two parameters: $\bar \omega_{yr}$ and $R_{yr}/l$, where $R_{yr}$ is the radius of the circle in Fig.~\ref{fig:robot}(b). If the sign of the angular velocity is flipped every $T$ seconds, the output can be captured by $\theta$ (Fig.~\ref{fig:robot}(b)) and effective speed $v$ (distance traveled along a straight line per unit time).

A future direction of research is to exploit the efficiency of the simulator to train a neural network that models the input - output relationship of this problem. That neural network then can serve as a look-up table (without performing any simulation) to formulate the control signal, given the prescribed trajectory.

\section{Control for path planning}
\label{sec:robotcontrol}

In this section, we present two examples of the inverse problem where the trajectory (circle and square) is prescribed and the angular velocity of the motor has to be computed. All the physical parameters are the same as those in Section~\ref{sec:method}: $\{ C_t, C_r, C_{yr}, l/R, l/r_0\} = {3.0, 2.8, 2.0, 6.875, 34.375}$. The intrinsic time-scale is $\mu_0 l^4/(EI)$ = 2.207 seconds. Number of tails is $N=2$.


In the first example in Fig.~\ref{fig:trajectory_design}(a), the robot starts from point $A$ and needs to follow a circular path (the radius of this circle is not equal to $R_{yr}$). Here, we introduce one of the simplest possible control schemes (Fig.~\ref{fig:trajectory_design}(c)) where the angular velocity of the motor is either $\omega_H$ or $\omega_L$ ($\omega_H = - \omega_L$). We rather arbitrarily choose $\omega_H=10$ (and $\omega_L=-10$). The remaining task is to compute the timepoints ($t_1, t_2, \ldots$ in Fig.~\ref{fig:trajectory_design}(c)) at which the angular velocity has to be switched.
To make the robot swim along a circle, the motor first rotates counterclockwise for normalized duration $t_1$, causing the robot to traverse a clockwise arc of angle $\theta$. Then, the motor rotates clockwise for a marginally shorter duration $t_2-t_1$, causing the robot to move through a slightly smaller arc of angle $\theta-\Delta \theta$. This input, alternating between a short counterclockwise rotation and a longer clockwise rotation, is repeated to form a zig-zag circular path in Fig. \ref{fig:trajectory_design} (c). 

In the second example in Fig.~\ref{fig:trajectory_design}(b), the robot has to follow a rectangular trajectory. It is obvious that the robot will follow a straight line if $\bar \omega$ switches between $\omega_H$ and $\omega_L$ every $T$ seconds. In Fig.~\ref{fig:trajectory_design}(d), this is the case when the robot has to follow a straight line ($0 \le \bar t \le t_C$, $t_D \le \bar t \le t_E$, $t_F \le \bar t \le t_G$, $t_H \le \bar t \le t_I$). Once the robot arrives at one corner of the rectangular path, $C$ as displayed in Fig. \ref{fig:trajectory_design}(b), the motor keeps rotating in one direction (time from $t_C$ to $t_D$) until the robot finishes turning $90^\circ$ and it reaches point $D$. The same protocol of turning is applied at points $E, G,$ and $I$.
%


\section{Conclusions and future work}
\label{sec:conclusion}

In summary, we built a framework comprised of a simple untethered soft robot, a numerical simulator, and a simple control scheme that enables the robot to follow any prescribed trajectory.
Our low-cost, easy-to-assemble, untethered soft flagellated robot offers a convenient and practical platform for users to study hydrodynamics near the air-liquid interface in viscous fluid. 
%
The robot is able to follow any prescribed 2D trajectory through a simple control method with a single binary input.
In addition to the low cost, this simplicity points to possible miniaturization of the robot.
As the size of the robot gets smaller, viscous effects start to dominate and the flow approaches low Reynolds number.
The propulsion mechanism of the proposed robot relies on low Reynolds assumption and provides a blueprint for micro-robots.



\section*{Acknowledgments}
We acknowledge support from the National Science Foundation (Award \# IIS - 1925360) and the Henry Samueli School of Engineering and Applied Science, University of California, Los Angeles.




\end{document}